\documentclass[letterpaper, 10 pt, conference]{ieeeconf}  

\IEEEoverridecommandlockouts                              

\usepackage{graphicx}
\usepackage[table]{xcolor}
\usepackage[caption=false,font=footnotesize]{subfig}
\usepackage[export]{adjustbox}
\usepackage{amssymb}  
\usepackage{siunitx}
\usepackage[nolist]{acronym}

\makeatletter
\let\NAT@parse\undefined
\makeatother
\usepackage[colorlinks,citecolor=red,urlcolor=blue,bookmarks=false,hypertexnames=true]{hyperref}

\usepackage[capitalise]{cleveref}
\usepackage[numbers]{natbib}
\usepackage{booktabs}
\usepackage{tabularx}
\usepackage{multirow}

\newcolumntype{C}{>{\centering\arraybackslash}X}
\newcolumntype{R}{>{\raggedleft\arraybackslash}X}

\begin{acronym}
    \acro{NC}[NC]{Neuromorphic Computing}
    \acro{NN}[NN]{Neural Network}
    \acro{ANN}[ANN]{Artificial Neural Network}
    \acro{CNN}[CNN]{Convolutional Neural Network}
    \acro{SNN}[SNN]{Spiking Neural Network}
    \acro{ReLU}[ReLU]{Rectified Linear Unit}
    \acro{ROI}[ROI]{Region of Interest}
    \acro{MSE}[MSE]{Mean Squared Error}
    \acro{IF}[IF]{Integrate-and-Fire}
\end{acronym}

\title{\LARGE \bf
    Detection of Fast-Moving Objects with Neuromorphic Hardware
}

\author{Andreas Ziegler$^{1}$, Karl Vetter$^{1}$, Thomas Gossard$^{1}$, Jonas Tebbe$^{1}$, Sebastian Otte$^{2}$, and Andreas Zell$^{1}$
    \thanks{$^{1}$Andreas Ziegler, Karl Vetter, Thomas Gossard, Jonas Tebbe, and Andreas Zell are with the University of Tübingen.%
            $^{2}$Sebastian Otte is with the University of Lübeck.
            Corresponding author {\tt\small andreas.ziegler@uni-tuebingen.de}.\newline
    }
    \thanks{This research was partially funded by Sony AI.}
}

\begin{document}
    \maketitle
    \thispagestyle{empty}
    \pagestyle{empty}
    \begin{abstract}
        \ac{NC} and \acp{SNN} in particular are often viewed as the next generation of \acp{NN}.
        \ac{NC} is a novel bio-inspired paradigm for energy efficient neural computation, often relying on \acp{SNN} in which neurons communicate via spikes in a sparse, event-based manner.
        This communication via spikes can be exploited by neuromorphic hardware implementations very effectively and results in a drastic reductions of power consumption and latency in contrast to regular GPU-based \acp{NN}.
        In recent years, neuromorphic hardware has become more accessible, and the support of learning frameworks has improved.
        However, available hardware is partially still experimental, and it is not transparent what these solutions are effectively capable of, how they integrate into real-world robotics applications, and how they realistically benefit energy efficiency and latency.
        In this work, we provide the robotics research community with an overview of what is possible with \acp{SNN} on neuromorphic hardware focusing on real-time processing.
        We introduce a benchmark of three popular neuromorphic hardware devices for the task of event-based object detection.
        Moreover, we show that an \ac{SNN} on a neuromorphic hardware is able to run in a challenging table tennis robot setup in real-time.
    \end{abstract}    
    
    \section*{Supplementary Material}
    Additional resources are available at: \url{https://cogsys-tuebingen.github.io/snn-edge-benchmark}
    
    \section{INTRODUCTION}
    
    \acfp{SNN} mimic the spiking behavior of biological neurons, offering a biologically inspired approach to \acf{NN} computation.
    Unlike traditional artificial neurons that produce real-valued outputs, spiking neurons receive input spikes and integrate them in the state of the neuron, called membrane potential.
    \begin{figure}[!t]
        \centering
        \includegraphics[width=1.0\linewidth]{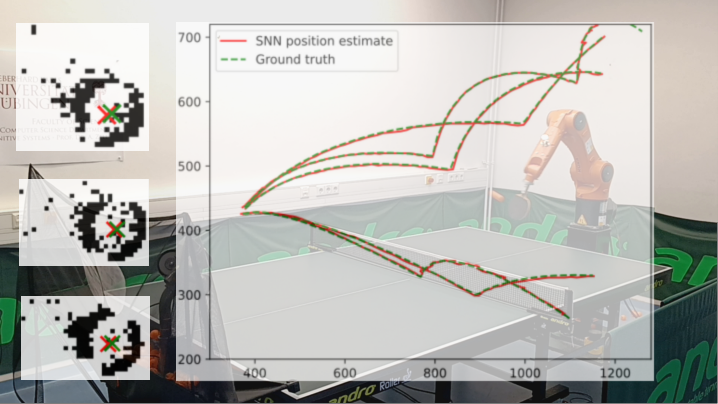}
        \caption{
                Left: Three examples of 2D ball detections in an accumulated event frame which serves as the input to the \acf{SNN} with {\color{green}ground truth in green} and the {\color{red}estimated position in red}.
                Right: Five observed 2D trajectories in the camera frame of the event-based camera with {\color{green}ground truth in green} and the {\color{red}estimated positions in red}.
                Background: The table tennis robot setup with the robot hitting back a table tennis ball in a rally.
                }
        \label{fig:eyecatcher}
    \end{figure}
    When the neuron's membrane potential reaches a defined threshold, the neuron emits a spike that propagates through the network, and resets its membrane potential.
    More theoretical work on \acp{SNN} has already been present in research for decades~\cite{Izhikevich2003tnn}\cite{Paugam2012computing}\cite{Bellec2018anips}\cite{Higuchi2024icml}.
    With the availability of more computing power through GPUs, it became feasible to train and simulate \acp{SNN} for real-world tasks.
    However, simulating \acp{SNN} is very inefficient on GPUs since the \acf{NC} paradigm is fundamentally different.
    Therefore, running \acp{SNN} on GPUs, though possible, is not a reasonable option for real-world applications, especially not for real-time robotics applications.
    
    In contrast, neuromorphic hardware, specifically designed for efficient \ac{SNN} processing, can leverage the sparsity and binary nature of \ac{SNN} outputs, resulting in a drastic reduction of power consumption and latency compared to conventional \acp{NN}.
    
    But the efficient processing of \acp{SNN} relies on gathering suitable data.
    Conventional cameras capture frames at a fixed frame rate, providing information about brightness and color.
    Event-based cameras, on the other hand, report asynchronous brightness changes per pixel, without measuring the absolute brightness~\cite{Lichtsteiner2008jssc}\cite{Gallego2020pami}.
    These cameras offer a high dynamic range, a temporal resolution in the order of $\mu$s, as well as energy and data efficiency.
    The binary output of event-based cameras aligns with the spike format of \acp{SNN}, making them a good match.

    This synergy between event-based cameras and \acp{SNN} presents an opportunity to improve real-time performance in robotics.
    One such area where these advancements can be applied is table tennis robotics, which has gained popularity in recent years~\cite{Ziegler2023corlw}.
    While not yet able to compete with professional players, table tennis robots are an exciting research environment to bring algorithms towards their limits.
    We thus use a table tennis robot scenario as a benchmark suite for event-based neuromorphic perception systems.
    
    A primary perception task for a table tennis robot system is fast and accurate ball detection.
    So far, most research uses frame-based cameras together with a \acf{CNN} based ball detection or a classical computer vision approach~\cite{Tebbe2019gcpr}\cite{DAmbrosio2023rss}\cite{GomezGonzalez2019robotics}\cite{Ding2022iros}.
    While ball detection solutions with frame-based cameras are successfully used, the high temporal resolution of event-based cameras promises faster and more frequent ball detections.
    This can improve the prediction of the ball's trajectory, which allows a faster robot control.
    As mentioned, \acp{SNN} align well with event data and can handle more complex scenarios, like a cluttered environment, compared to model-based solutions.
    
    This work explores the combination of an event-based camera and \acp{SNN} for object detection and their deployment on neuromorphic edge devices.
    We analyze the potential benefits of this fusion, and discuss the limitations of a deployment on neuromorphic edge devices.
    We report the error and run-time, both in simulation and on multiple neuromorphic edge devices, namely the DynapCNN\footnote{https://www.synsense.ai/products/dynap-cnn/} from SynSense, Akida\footnote{https://brainchip.com/akida-enablement-platforms/} from BrainChip and Loihi2\footnote{https://www.intel.com/content/www/us/en/research/neuromorphic-computing-loihi-2-technology-brief.html} from Intel.
    Benchmarking these three widely used neuromorphic edge devices on our use case of object detection gives an example of what is possible with these edge devices in robotics perception applications. 
    
    In summary, our \textbf{contributions} with this work are:
    \begin{itemize}
        \item We demonstrate the effectiveness of \acp{SNN} for event-based object detection
        \item We conduct a comparative study of available neuromorphic edge devices
        \item We present a streamlined integration of event-based sensing, neuromorphic processing, and robot arm planning and control in a table tennis robot setup
        \item We provide a publicly available benchmark dataset for event-based ball detection
    \end{itemize}
    
    \section{RELATED WORK}\label{sec:related_work}
    
    We will start with an introduction to event-based cameras in \cref{subsec:rw_even-cameras}.
    \acp{SNN} will be covered in \cref{subsec:rw_snn}.
    Different ways to train an \ac{SNN} are covered in \cref{subsec:rw_ann_to_snn} and in \cref{subsec:rw_snn_train}.
    We will finish the related work covering spiking object detection in \cref{subsec:rw_s_object_detection}.
    
    \subsection{Event-Based Cameras}\label{subsec:rw_even-cameras}
    
    Event-based cameras, also known as neuromorphic cameras or dynamic vision sensors, have gained considerable attention in computer vision and robotics due to their unique characteristics and advantages over traditional frame-based cameras~\cite{Lichtsteiner2008jssc}\cite{Gallego2020pami}.
    Event-based cameras operate by detecting logarithmic changes in brightness asynchronously on a per-pixel basis, reporting events only when the brightness change exceeds a specified threshold.
    
    The asynchronous event-based nature of event-based cameras enables high temporal resolution and low latency, making them particularly suitable for capturing fast-moving objects~\cite{Monforte2020aicas}\cite{Mitrokhin2018iros}\cite{Forrai2023icra} and scenes with a high dynamic range~\cite{Perot2020neurips}\cite{Stoffregen2020eccv}.
    
    \subsection{Spiking Neural Networks (SNN)}\label{subsec:rw_snn}
    
    \acp{SNN} have garnered substantial interest in the field of \acp{NN} due to their biological inspiration and potential for energy-efficient computation~\cite{Lemaire2023lncs}.
    \acp{SNN} differ from traditional \acfp{ANN} by modeling the spiking behavior of biological neurons more closely.
    They rely on sparse and binary spikes for information processing.
    \acp{SNN} can represent different neuron models, such as the \ac{IF}~\cite{Brunel2007bc} or the leaky \ac{IF}~\cite{Lu2022fnins} model.
    
    Studies indicate that \acp{SNN} are more energy-efficient~\cite{Lemaire2023lncs} and excel in processing spatio-temporal data due to their use of sparse, binary signals~\cite{Blouw2019nice}.
    Consequently, \acp{SNN} are ideal for energy-sensitive hardware applications~\cite{Davies2018ieem}\cite{Akopyan2015tcad}.
    
    A significant advancement in \ac{SNN} research is the development of efficient training techniques.
    Converting pre-trained \acp{ANN} into \acp{SNN} will be described in \cref{subsec:rw_ann_to_snn} and direct training of \acp{SNN} using spike-based learning rules, in \cref{subsec:rw_snn_train}.
    
    \subsection{\acf{ANN} to \ac{SNN} Conversion}\label{subsec:rw_ann_to_snn}
    
    One approach to train \acp{SNN} involves converting pre-trained \acp{ANN} into \acp{SNN}~\cite{Cao2014ijcv}.
    This can be done by approximating the output of the \ac{ReLU} non-linearity with rate-code~\cite{Diehl2015ijcnn}.
    Rate-code represents the relative frequency of spikes, obtained by replacing the \ac{ReLU} with the Heaviside step function and setting a spiking threshold of one.
    This approximation captures the essential characteristics of the \ac{ReLU}, for output activities between zero and one.
    \ac{ReLU} outputs larger than one can not be represented by normal \ac{IF} neurons, as they can at most spike once per time step.
    However, a multi-spike neuron model~\cite{Weidel2021arxiv} allows the direct representation of the entire \ac{ReLU} output space through rate-code.
    
    \subsection{Direct \acf{SNN} Training}\label{subsec:rw_snn_train}
    
    There are two major lines of research to directly train \acp{SNN}.
    One is to train the model through biologically plausible local learning rules, e.g., Spike Timing Dependent Plasticity (STDP)~\cite{Diehl2015fncom}\cite{Bi1998jons}, the other one to use so-called surrogate gradients~\cite{Neftci2019spm}.
    Since STDP is not available for most of the \ac{SNN} frameworks used in this work, we will focus on surrogate gradients.
    
    Back-propagation in \acp{SNN} is akin to a specific instance of back-propagation through time (BPTT), typical for RNN training.
    However, the Heaviside function lacks differentiability at the threshold, and therefore, surrogate gradients are necessary for approximating gradients~\cite{Neftci2019spm}.
    
    \subsection{Spiking Object Detection}\label{subsec:rw_s_object_detection}
    
    Object detection is a fundamental task in computer vision, and traditional approaches typically use \acp{CNN}.
    In recent years, there has been growing interest in exploring \acp{SNN} for object detection tasks due to their potential for improved energy efficiency and event-driven processing~\cite{Davies2018ieem}\cite{Akopyan2015tcad}.
    
    The YOLO architecture~\cite{Redmon2016cvpr}, a popular object detection framework, has been successfully converted into a \ac{SNN} variant called Spiking-YOLO in~\cite{Kim2020aaai}.
    
    \acp{SNN} were also explored for specific object detection tasks.
    In~\cite{Debat2021ficn}, \acp{SNN} were employed to predict ball trajectories using data from event-based cameras.
    Their approach utilized spatio-temporal filters with weights incorporating delays to capture temporal information.
    They also employed leaky \ac{IF} neurons and trained the network using STDP.
    
    An \ac{SNN} was utilized in~\cite{Jiang2019fin} for detecting balls and pipes in a simulated environment.
    They used simulated neuromorphic data and designed an \ac{SNN} architecture specifically tailored for detecting circles, ellipses, and lines using the Hough transform.
    Their approach demonstrated successful detection using the designed \ac{SNN} without training, relying on hard-coded algorithms.
    
    Although promising, \acp{SNN} for object detection still faces challenges, including higher error rates compared to \ac{CNN}-based detectors.
    Further advancements in network architectures, training algorithms, spike-based encoding schemes, and hardware are necessary to bridge this performance gap.
    Moreover, optimizing the computational and memory efficiency of \ac{SNN} models is crucial for enabling real-time object detection on resource-constrained devices.

    \section{METHOD}\label{sec:method}
    
    We recorded a ball detection dataset to train our \acp{SNN}, described in \cref{subsec:data}.
    In this work, we used three state-of-the-art \ac{SNN} frameworks and a corresponding neuromorphic edge device.
    For each of them, we designed an \ac{SNN} architecture, conforming with their specific constraints, described in \cref{subsec:network_architecture}.
    The training with the three \ac{SNN} frameworks is covered in \cref{subsec:network_training}.
    
    \subsection{Data}\label{subsec:data}
    
    We describe the recording setup in \cref{subsubsec:recording_setup}, followed by an explanation of how the training labels were generated in \cref{subsubsec:data_generation}.
    The specifications of the dataset are summarized in \cref{tab:data}.
    \begin{table}[b!]
        \caption{
            Dataset specifications
        }
        \resizebox{\linewidth}{!}{
            \begin{tabular}{l|l}
                \hline
                Number of samples & Training: $8630$, validation: $531$, test: $531$ \\
                \hline
                Labeled samples   & Automatic labeled: $7569$, manually labeled: $2123$ \\
                \hline
                Cameras used      & Event-based: $2$x Prophesee EVK4 ($1280$x$720$) \\
                                  & Frame-based: $4$x FLIR Chameleon3 ($1280$x$1024$, $140$fps) \\
                \hline
                Event accumulation time & $1$ms \\
                \hline
                Ballgun           & Butterfly Amicus Prime, ball speed $4$m/s \\
                \hline
            \end{tabular}
        }
        \label{tab:data}
    \end{table}
    
    \subsubsection{Recording setup}\label{subsubsec:recording_setup}
    
\begin{figure}[b!]
        \centering
        \includegraphics[width=1.0\linewidth]{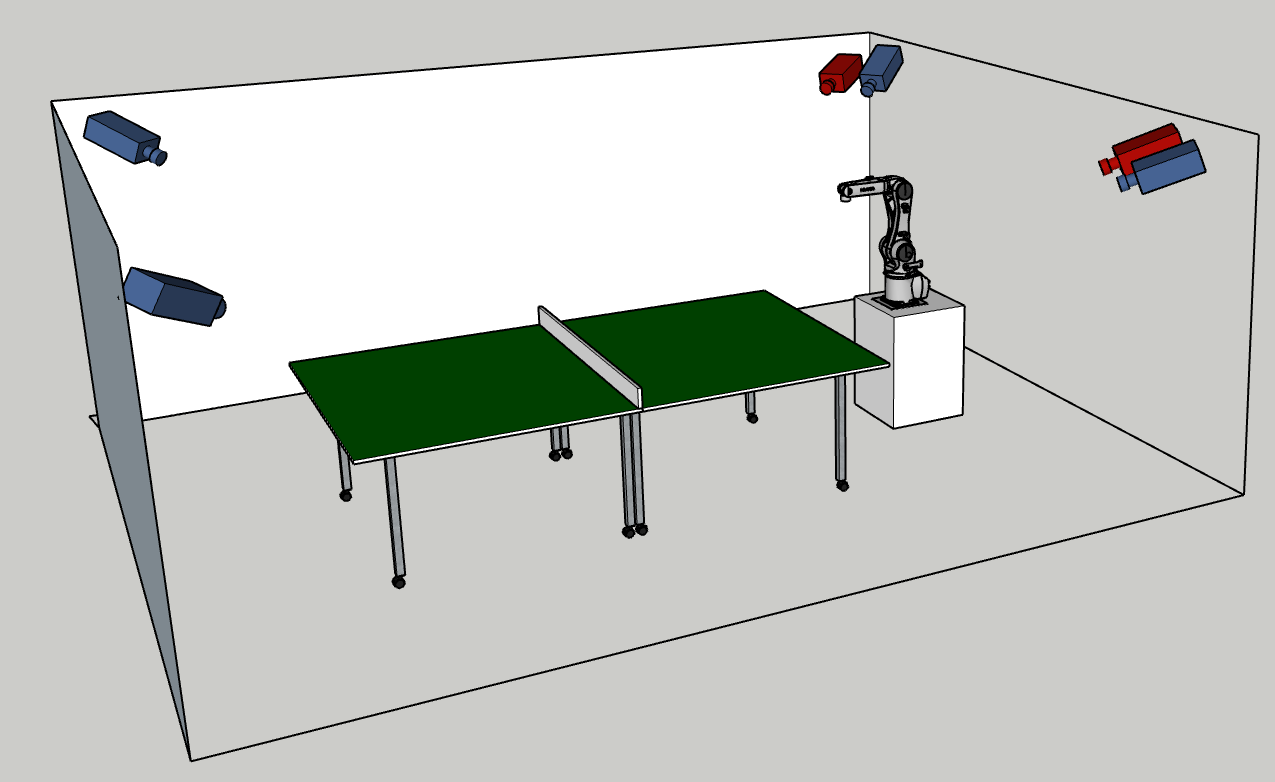}
        \caption{
            Our camera setup consisting of {\color{blue}four frame-based cameras (in blue)} and {\color{red}two event-based cameras (in red)} with baselines of $3$m to $5$m.
            Schematic is up to scale.
        }
        \label{fig:camera_setup}
    \end{figure}

    We used a table tennis robot setup as introduced in~\cite{Tebbe2019gcpr}.
    The setup was extended with two event-based cameras as described in~\cite{Ziegler2023corlw}.
    This camera system consists of four FLIR Chameleon3 frame-based cameras (140 fps, 1280x1024 pixels) and two Prophesee EVK4 event-based cameras (1280x720 pixels).
    Camera bias settings for the event-based camera were configured to minimize noise, and so that most events would be caused by the flying ball.
    The whole table tennis robot system is visualized in \cref{fig:camera_setup}.
    To calibrate this camera system containing frame- and event-based cameras, we used the wand-based calibration approach introduced in~\cite{Gossard2023icra}.
    The frame-based cameras are used to detect and triangulate the position of the table tennis ball in 3D, as described in~\cite{Tebbe2019gcpr}.
    The event stream from the event-based cameras is used as input for our \ac{SNN} approach.
    
    A Butterfly Amicus Prime ball gun, with default speed settings ($4$m/s), was used to shoot balls, as shown in the background of~\cref{fig:eyecatcher}.
    
    \subsubsection{Data generation}\label{subsubsec:data_generation}
    
    The input of our \acp{SNN} is a matrix of size $64$x$64$ pixels, where over a time range of $1$ms the pixel value is set to one if at least one event occurred at the pixel and zero otherwise.
    To reduce computation and achieve faster inference times, we favored smaller networks and thus did not include the polarity information of the events.
    As in~\cite{Vasco2016iros}, our event representation does also not differentiate between pixels with more or less events, as long as they have any.
    This reduction in information was carried out as it led to better performance in our experiments.
    A dynamic \ac{ROI} was used to crop $64$x$64$ pixels out of the $1280$x$720$ pixels output of the event-based camera, similar to~\cite{Tebbe2019gcpr}.
    This was done by setting the last known ball position as the center of the \ac{ROI}.
    This has the benefit of being applicable in real-time using the system's own past ball position estimates.
    Since the system starts each trajectory without past position estimates, we used manually defined start regions.
    This limitation could be alleviated by using the position of a frame-based detection system or an \ac{CNN} based ball detector for the initialization.
    
    The output of the \acp{SNN} is the 2D pixel position of the detected ball.
    Therefore, the required ground truth is the 2D ball position in the image frame of the event-based camera.
    We used two ways to generate the 2D ball positions, serving as ground truth.
    We projected the 3D ball positions from the frame-based camera system into the camera frame of the event-based cameras.
    This way, we generated $7569$ ground truth positions.
    Since we only obtain 3D positions every $7$ms, we additionally labeled the 2D ball position manually on an additional set of $2123$ event frames.
    
    \subsection{Network Architecture}\label{subsec:network_architecture}

    The state-of-the-art \ac{SNN} frameworks and their corresponding neuromorphic edge device, we used in this work, each have their constraints.
    Some constraints are the input/output resolutions per layer, the number of output weights per neuron, and the type of supported layers.
    These constraints make it difficult to run any more complex object detection network.
    Therefore, the challenge was to find a trade-off between the network's complexity and the constraints and limitations of the neuromorphic edge devices. 
    
\begin{figure}[t!]
        \centering
        \includegraphics[width=0.75\linewidth]{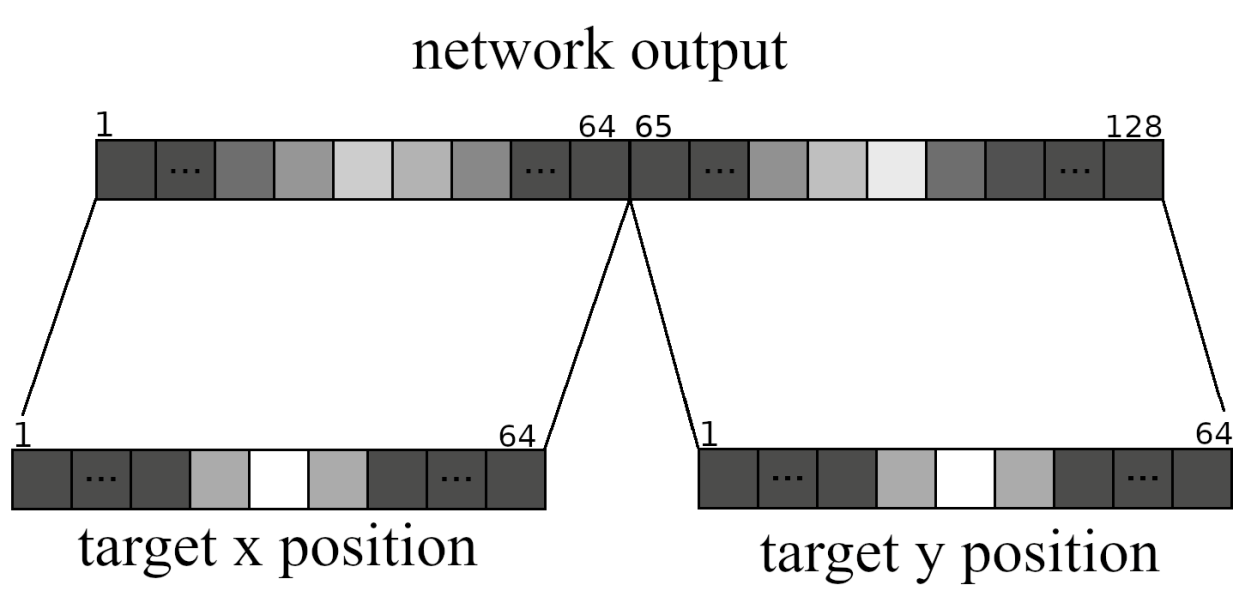}
        \caption{
            The network has $128$ output neurons, which are split into two populations.
            Each neuron in the first population represents an $x$-position, and each neuron in the second represents a $y$-position.
            The target is also split into an $x$- and $y$-target, each setting one as the target activity for the correct neuron, $0.5$ for the two adjacent neurons, and zero for all others.
            Values are represented using brightness, with larger values being brighter.
        }
        \label{fig:classification_output}
    \end{figure}

    We cast the task of detecting the table tennis ball in 2D as a classification task.
    This allows fewer time steps and, therefore, a faster inference time than a regression network due to the rate-code approximation.
    We treat the $x$- and $y$-positions as two independent classification tasks.
    In this case, the ball's $x$-position can be one of $64$ classes, as can the $y$-position.
    A visualization of the network's output is shown in \cref{fig:classification_output}.
    Although the neuromorphic edge devices support higher input resolutions, we decided to use only $64$x$64$ pixels to improve the inference time, which is crucial for our real-time use case.
    To receive the network's prediction, the neurons with the maximal output are determined in each of the two populations.
    
    The network consists of four layers, the first two being convolutional layers and the last two being linear layers.
    A description of the network is provided in \cref{tab:network_architecture}.
    The activation function and the pooling are given in the following order: sinabs / MetaTF / Lava.
    “None” indicates that no layer is used for the corresponding \ac{SNN} framework.
    \begin{table}[t!]
        \caption{
            Network architecture for the three \ac{SNN} frameworks.
            The details are given in the following order: sinabs / MetaTF / Lava.
            “None” indicates that no layer is used for the corresponding \ac{SNN} framework.
            Where stride is not explicit, it defaults to $1$.
        }
        \begin{tabularx}{\linewidth}{c|C}
            Layer &  Layer Specifics\\
            \hline
            \hline

            \multirow{3}{*}{1} &  ConvLayer(outChannels = 4, kernelsize = 5x5, stride = 2)\\
            &  Multispike IF / (BatchNorm \& QuantizedReLU) / LIF\\
            &  AveragePool(2x2) / MaxPool(2x2 / None\\
            \hline
            \multirow{3}{*}{2} &   ConvLayer(outChannels = 4, kernelsize = 3x3)\\
            &  Multispike IF / (BatchNorm \& QuantizedReLU) / LIF\\
            &  AveragePool(2x2) / None / None\\
            \hline
            \multirow{2}{*}{3} &  LinearLayer(outChannels = 64)\\
            &  Multispike IF / (BatchNorm \& QuantizedReLU) / LIF\\
            \hline
            \multirow{2}{*}{4} &  LinearLayer(outChannels = 128)\\
            &  Multispike IF / None / LIF\\
        \end{tabularx}
        \label{tab:network_architecture}
    \end{table}
    Since all the \ac{SNN} frameworks we used had different limitations, we could not use one \ac{SNN} architecture for all of them.
    Therefore, we modified the \ac{SNN} to comply with the constraints of the corresponding framework.
    
    \subsubsection{sinabs (DynapCNN)}
    While the \ac{SNN} trained on the GPU processes frames and is clocked, the \ac{SNN} on the DynapCNN processes events and is asynchronous.
    This difference leads to a gap between the two \acp{SNN}.
    We use the multi-spike learning approach~\cite{Weidel2021arxiv} supported by sinabs, since it reduces this gap.
    The multispike \ac{IF} neurons are followed by average pooling to further reduce network size and remain within the limits for the number of neurons per layer imposed by the DynapCNN.
    
    \subsubsection{MetaTF (Akida)}
    
    The network we used to train with the MetaTF framework uses batch normalization layers in all layers except the last one, as it led to improved performance, and Akida supports the use of biases, which are necessary to integrate batch normalization into an \ac{SNN} without significant effect on the inference time.
    Quantized ReLUs are used because the Akida chip squashes the rate-code approximation of the ReLU into one time step, where it is then represented by a step-wise quantized ReLU.
    In the first layer, we use max pooling instead of average pooling, since this is the pooling type supported by the MetaTF framework.
    For the use on the chip, pooling comes with additional restrictions, which is why there is no second pooling layer and instead a stride of two in the second convolutional layer.
    
    \subsubsection{Lava (Loihi2)}
    
    For the network trained with Lava, we use leaky \ac{IF} neurons, since they are the standard neuron type supported by the Loihi2 chip and performed better than \ac{IF} neurons.
    An increased number of channels (8, 16, 64, and 128) is used, as this was necessary to reach comparable accuracy.
    No pooling layers were used, as they cannot be mapped onto the Loihi2 chip.
    
    \subsection{Training}\label{subsec:network_training}
    
    As mentioned in \cref{subsec:rw_snn}, there are two ways to train \acp{SNN}.
    The conversion from an \ac{ANN} to an \ac{SNN} and direct \ac{SNN} training.
    The former approach has notable drawbacks, including a typical decline in accuracy, especially when using few time steps, and a high spike rate~\cite{Eshraghian2023ieee}.
    Therefore, we used direct \ac{SNN} training for the DynapCNN and Intel's bootstrap method for Loihi2.
    Akida uses an equivalent representation of an \ac{SNN} using a  step-wise quantized ReLU, which can be trained using quantization-aware training and allows for processing in a single time step.

    While the standard loss for a classification task would be the cross entropy loss, the \ac{MSE} proved more suitable for our use case.
    We observed that the cross entropy loss leads to \acp{SNN} with very large activations.
    Large activations result in more spikes, causing more synaptic operations, making the network slower when used on edge devices.
    Using the \ac{MSE} requires a target output of the same shape as the network output.
    The target that worked best was setting the correct neuron's target output to one, the target output for the two neurons directly adjacent to $0.5$, and all other neuron targets to $0$, as visualized in \cref{fig:classification_output}.
    
    For all \ac{SNN} frameworks, we trained the networks on a GPU and deployed them on the edge devices for inference.
    We used all the $7569$ automatically generated 2D ground truth positions and additional $1061$ manually labeled positions (as described in \cref{subsubsec:data_generation}) for the training set and $531$ manually labeled 2D positions for the validation and the test set.
    Since the training depends on the \ac{SNN} framework, we describe the training for each one.
    Common hyperparameters are listed in \cref{tab:training_hyperparameters}.
    \begin{table}[t!]
        \centering
        \caption{
            Common Hyperparameters of the \ac{SNN} frameworks.
        }
            \begin{tabular}{l|l|l|l}
                \ac{SNN} framework & Learning rate & Batch size & Optimizer\\
                \hline
                \hline
                sinnabs (DynapCNN) & $0.0001$ & $200$ & Adam \\
                \hline
                Lava (Loihi2)      & $0.001$ & $100$ & Adam  \\
                \hline
                MetaTF (Akida) & $0.0001$ & $1000$ & Adam  \\
            \end{tabular}
        \label{tab:training_hyperparameters}
    \end{table}
    
    \subsubsection{Training with sinabs (DynapCNN)}\label{subsubsec:train_sinabs}
    
    To train \acp{SNN} for the  DynapCNN, SynSense provides sinabs\footnote{https://github.com/synsense/sinabs} as a framework.
    We use the multi-spike learning approach~\cite{Weidel2021arxiv} supported by sinabs, since it closely matches the neural behavior on the chip.
    The sinabs framework additionally allows for the tracking of the network's synaptic operations, which can then be used as an additional loss term to prevent over activated neurons on the DynapCNN.
    With the multi-spike approach, the number of generated spikes when the membrane potential exceeds the threshold is proportional to the membrane potential.
    
    To minimize the quantization loss, the distribution of weights needs to be such that no extreme outlier weights exist, which would decrease the quantization accuracy for all other weights.
    To normalize the weights on the DynapCNN, we used an additional loss term using only the absolute maximum of the weights for every layer.
    The \ac{SNN} was trained using the periodic exponential function as surrogate gradient function  \cite{Neftci2019spm} and backprop through time.

    \subsubsection{Training with MetaTF (Akida)}\label{subsubsec:train_akida}
    
    To train \acp{SNN} for the Akida, BrainChip provides MetaTF\footnote{https://doc.brainchipinc.com/index.html} as a framework.
    MetaTF directly trains an \ac{ANN} approximating a \ac{SNN} through quantization.
    We followed the guidelines of the MetaTF documentation to train the \ac{SNN} with MetaTF.
    
    \subsubsection{Training with Lava (Loihi2)}\label{subsubsec:train_lava}
    
    To train \acp{SNN} for the Loihi2, Intel provides Lava\footnote{https://lava-nc.org/index.html} as a framework.
    Direct training is supported by Lava with an enhanced version of the SLAYER framework~\cite{Shrestha2018neurips}.
    We followed the guidelines of the Lava documentation to train the \ac{SNN} with Lava-dl.
    The neuron thresholds were set to $0.25$ and voltage decay to $0.05$.
    
    \section{EXPERIMENTS}\label{sec:experiments}
    
    In the first experiment, described in \cref{subsec:experiment_offline}, we evaluate the detection error and measure the time per forward pass of the three different \ac{SNN} frameworks in simulation as well as on the neuromorphic edge device.
    \Cref{subsec:experiment_online} describes the second experiment, in which we run the \ac{SNN} on a neuromorphic edge device within a table tennis robot setup in real-time.

    \begin{figure}[t!]
        \centering
        \includegraphics[width=1.0\linewidth]{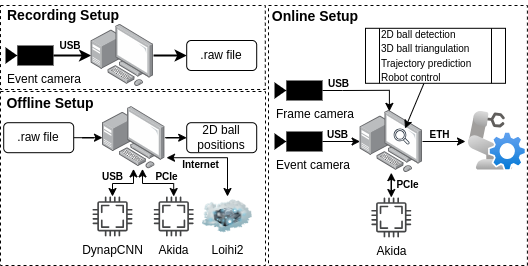}
        \caption{
            Left: Experiment setup for the offline experiment.
            Right: Experiment setup for the online experiment.
        }
        \label{fig:experiment_setup}
    \end{figure}
    
    \subsection{Offline experiment}\label{subsec:experiment_offline}
    
    In this experiment, we compare the different \ac{SNN} frameworks in terms of error and time per forward pass, given our robotic perception task.
    Using our recorded data, we ran the 2D ball detection $10$ times for BrainChip and sinabs and $5$ times for Lava.
    The setup is depicted in \cref{fig:experiment_setup} on the left.
    Five recorded 2D ball trajectories and three 2D ball detections are shown in \cref{fig:eyecatcher}.
    
    The error, as well as the time per forward pass, was recorded for every run, and we report the mean and standard deviation in \cref{tab:offline_results}.
    \begin{table}[t!]
        \caption{
            Ball center error and time per forward pass of the different \ac{SNN} frameworks and on their corresponding neuromorphic edge device.
            Steps indicate how many time steps were used for one inference pass.
            The results are reported with mean and std. dev over multiple runs.}
        \resizebox{\linewidth}{!}{
            \begin{tabular}{l|c|l|l|S[table-format=1.2(2)]|S[table-format=4.2(2)]}
                Framework & Sim & Hardware & Steps & {Error [pixel]} &  {One forward pass [ms]} \\
                \hline
                \hline
                \rowcolor{gray!20}BrainChip & \checkmark &        & 15 (1) &  \multicolumn{1}{c|}{\textbf{1.44 $\pm$ 1.41}} & \multicolumn{1}{c}{1.75 $\pm$ 0.82}       \\
                \hline
                \rowcolor{gray!20}sinabs    & \checkmark &        &  8 (1) &  \multicolumn{1}{c|}{1.55 $\pm$ 1.89}  & \multicolumn{1}{c}{\textbf{1.59 $\pm$ 1.61}}       \\
                \hline
                \rowcolor{gray!20}Lava      & \checkmark &        & 20     &  \multicolumn{1}{c|}{1.57 $\pm$ 1.55}  & \multicolumn{1}{c}{1764.89 $\pm$ 1374.36} \\
                \hline
                BrainChip & & Akida    & 15 (1) &  \multicolumn{1}{c|}{\textbf{1.44 $\pm$ 1.41}} & \multicolumn{1}{c}{2.20 $\pm$ 0.35}       \\
                \hline
                sinabs    & & DynapCNN &  8     &  \multicolumn{1}{c|}{1.59 $\pm$ 1.83}  & \multicolumn{1}{c}{46.04 $\pm$ 21.38}     \\
                \hline
                Lava      & & Loihi2   & 20     &  \multicolumn{1}{c|}{1.57 $\pm$ 1.55}  & \multicolumn{1}{c}{1458.57 $\pm$ 230.500}   \\
                \hline
                TensorFlow & & RTX 2080 Ti & - & \multicolumn{1}{c|}{3.55 $\pm$ 1.49} & \multicolumn{1}{c}{3.49 $\pm$ 0.62}
            \end{tabular}
        }
        \label{tab:offline_results}
    \end{table}
    Steps indicate how many time steps were used for the rate-code for one inference pass, a trade-off between speed and accuracy.
    The steps count in brackets represents the number of time steps when using multispike neurons.
    We also report the results of an existing setup~\cite{Tebbe2021Tuebingen}, based on~\cite{GomezGonzalez2019robotics}, as a comparison.
    This setup uses a frame-based camera and a \ac{CNN} with similar image resolution and number of convolutional layers on an RTX 2080 Ti GPU.
    
    As can be seen in \cref{tab:offline_results} the BrainChip framework has the lowest ball center error of $1.44$ pixels.
    The error of sinabs and Lava are $1.55$ pixels and $1.57$ pixels, respectively, quite close together.
    
    Important to note is that depending on the neuromorphic edge device, the total time per forward pass is heavily influenced by the integration into the system.
    The total time for a forward pass includes the inference time and, depending on the neuromorphic edge device data transfer, pre- and post-processing.
    Therefore, we list the total time per forward pass, the inference time, and the power consumption in \cref{tab:timings}.
    \begin{table}[t!]
        \vspace{0.25cm}
        \caption{
            Total time per forward pass and inference time of the different neuromorphic edge devices with mean and std. dev over $10$ runs and the power consumption.
            $^\dagger$Processing one input on the DynapCNN takes an average of $1.765$ms, but with delays from the USB connection, the time for a forward pass is significantly longer.
            $^*$Taken from specs.
        }
        \resizebox{\linewidth}{!}{
            \begin{tabular}{l|S[table-format=4.2(2)]|S[table-format=1.2(2)]|l}
                Device & {One forward pass [ms]} & {Inference time [ms]} & Power consumption [mW] \\
                \hline
                \hline
                Akida & \multicolumn{1}{c|}{\textbf{2.20 $\pm$ 0.35}} & \multicolumn{1}{c|}{0.89 $\pm$ 0.28} & $\sim 4.5$\\
                \hline
                DynapCNN    & \multicolumn{1}{c|}{46.04 $\pm$ 1.38$^{\dagger}$} & \multicolumn{1}{c|}{0.82 $\pm$ 0.24} & $\sim 5^*$\\
                \hline
                Loihi2      & \multicolumn{1}{c|}{1458.57 $\pm$ 230.50} & \multicolumn{1}{c|}{\textbf{0.62 $\pm$ 0.01}} & $\sim 100^*$\\
                \hline
                RTX 2080 Ti & \multicolumn{1}{c|}{3.49 $\pm$ 0.62} & \multicolumn{1}{c|}{1.77 $\pm$ 0.28} & $\sim 250000^*$
            \end{tabular}
        }
        \label{tab:timings}
    \end{table}
    
    For the BrainChip Akida, which is installed as a PCIe card, the time to load data onto the device and return the results is a relatively small overhead ($0.89$ms inference time and $2.20$ms for a forward pass).
    
    The DynapCNN from SynSense is connected to the PC via USB.
    We operate the DynapCNN in a streaming mode.
    This allows a fast inference time of $0.82$ms but does come with a relatively high delay coming from the USB connection and other sources, which leads to $46.04$ms for a forward pass.
    The streaming mode allows for a high data throughput with an average processing time of $1.765$ms on the chip but does not remove the delay caused by the USB connection.
    Using the DynapCNN with another event-based camera might improve the setup and therefore, remove the mentioned delays.
    The recent development of combining an event-based camera with an \ac{SNN} on the same chip, as the Speck\footnote{https://www.synsense.ai/products/speck-2/} from SynSense, looks promising.
    However, the current camera resolution of $128$x$128$ pixels is too low for our application.
    
    For Loihi2, we did not have direct access to an edge device, but only through a virtual machine provided by Intel via the Intel Neuromorphic Research Community\footnote{https://intel-ncl.atlassian.net/wiki/spaces/INRC/overview}.
    With this Loihi2 setup, the time per forward pass is with $1458.57$ms two orders of magnitude slower than the other devices.
    This is mainly due to inefficient communication between the virtual machine and the Loihi2 chip, which is a consequence of our current implementation not being well optimized in this regard.
    Note that the inference time on the GPU is longer than on all neuromorphic edge devices, however, the good hardware integration of the GPU shows its advantage.
    
    Regarding the power consumption, we measured the one for Akida, but had to rely on specs for the other devices.
    The power consumption for Akida and DynapCNN is quite close together.
    Loihi2, being a much more powerful and versatile platform, consumes an order of magnitude more energy, still several orders of magnitude lower than a GPU, showing the benefits of \acf{NC} on neuromorphic edge devices for robotic applications.
    
    Overall, we can see that while the errors of the different neuromorphic frameworks are similar, the difference in inference time and the time per forward pass varies significantly.
    As already mentioned, the way in which the neuromorphic edge devices are connected to the PC plays a crucial role.
    
    \subsection{Online (real-time) experiment}\label{subsec:experiment_online}
    
    In this experiment, we integrated the \ac{SNN} based ball detection into our table tennis robot system, described in \cref{subsec:data}.
    As explained in~\cite{Tebbe2019gcpr}, to determine the 3D position of the table tennis ball, at least two cameras need to detect the ball in order to perform triangulation.
    Since we only have one device for each of the three \ac{SNN} frameworks, covered in this work, we used one event-based camera with the introduced \ac{SNN} ball detection together with a frame-based camera and the ball detection presented in~\cite{Tebbe2019gcpr}.
    We used the Akida PCIe card from BrainChip, since the system integration does not introduce latencies as high as the DynapCNN does, and we did not yet have access to a Loihi2 edge device.
    The experimental setup is depicted in \cref{fig:experiment_setup} on the right, and the physical setup is shown in the background of~\cref{fig:eyecatcher}.
    In this experiment, we shot 15 table tennis balls with a Butterfly Amicus ball gun.
    Since the 3D triangulation and robot arm control are not part of this work, we used the existing ones from~\cite{Tebbe2019gcpr}.
    We report the ball return rate of the robot, since controlling the landing point of the ball is out of scope for this work.
    Given this setup, we achieved a ball return rate of $1.0$.
    
    As previously mentioned, we used one event-based camera with the \ac{SNN} based ball detection and a frame-based camera with an existing ball detection to triangulate the 3D ball position.
    Although our online experiment does not rely solely on our \ac{SNN} based ball detection, we have shown that this would be possible, and future work could involve evaluating the system with an additional Akida PCIe card.
    
    \section{CONCLUSION}\label{sec:conculsion}

    With neuromorphic hardware more accessible in recent years, \acf{NC} and \acfp{SNN} in particular has become more relevant for robotics.
    In this work, we used an event camera in combination with \acp{SNN} for ball detection as a real-time perception task.
    Three different \ac{SNN} frameworks, namely sinabs, MetaTF, and Lava, and a corresponding neuromorphic edge device (DynapCNN, Akida, and Loihi2) were used, and their errors, time per forward pass, inference time, and power consumption were compared.
    Moreover, we show that an \ac{SNN} on a neuromorphic hardware is able to run in a challenging table tennis robot setup in real-time.

    Despite the promise of asynchronous processing with \acp{SNN}, the current edge devices face limitations, primarily attributed to hardware integration.
    Our results show that the better a neuromorphic edge device is connected to the main compute unit, e.g., as a PCIe card, the better the overall run-time. 
    
    This work aims to provide the robotic research community with insights into the possibilities and challenges of deploying \acp{SNN} on current neuromorphic edge devices for real-time robotic applications.

    \addtolength{\textheight}{-1.0cm}   
    


    
    

    
    \bibliographystyle{IEEEtran}
    \bibliography{bibliography}
    
\end{document}